\documentclass{article}

\usepackage[preprint]{neurips_2025}

\usepackage{tikz}
\PassOptionsToPackage{numbers, compress}{natbib}
\usepackage[utf8]{inputenc} 
\usepackage[T1]{fontenc}    
\usepackage{hyperref}       
\usepackage{url}            
\usepackage{booktabs}       
\usepackage{amsfonts}       
\usepackage{nicefrac}       
\usepackage{microtype}      
\usepackage{xcolor}         
\usepackage{amsmath}
\usepackage{algorithm}
\usepackage{algpseudocode}
\usepackage{graphicx}
\usepackage{hyperref}
\usepackage{cite}
\title{NeuroVoxel-LM: Language-Aligned 3D Perception via Dynamic Voxelization and Meta-Embedding
}

\author{%
  Shiyu Liu \\
  School of Electrical and Electronic Engineering\\
  Nanyang Technological University\\
  \texttt{E250115@e.ntu.edu.sg} \\
  \And
  Lianlei Shan \\
  School of Computer Science and Technology \\
  University of Chinese Academy of Sciences \\
  \texttt{shanlianlei18@mails.ucas.edu.cn} \\
}

\begin{document}

\maketitle

\begin{abstract}
Recent breakthroughs in visual language models (VLMs) and multimodal massive language models (MLLMs) have made significant contributions to the evolution of 3D scene perception towards language-driven cognition. When dealing with sparse large-scale point clouds, existing 3D large-scale language models frequently encounter issues such as slow feature extraction and low feature representation accuracy. To address these limitations, we offer NeuroVoxel-LM, a novel framework that integrates neural radiation field (NeRF) with dynamic resolution voxelization and lightweight meta-embedding techniques. Specifically, a Dynamic Resolution Multiscale Voxelisation (DR-MSV) technique is developed, which adaptively modifies voxel granularity based on structural and geometric complexity metrics, greatly lowering computing cost while retaining geometric accuracy. Furthermore, the Token-level Adaptive Pooling Lightweight Meta-Embedding (TAP-LME) method is developed, which improves global representation through attentional weighting and residual fusion, resulting in a more refined semantic comprehension of NeRF weights. Systematic experiments demonstrate that the DR-MSV method improves point cloud feature extraction while maintaining geometric reconstruction accuracy, and that the TAP-LME method outperforms the maximum pooling-only method.
\end{abstract}

\section{Introduction}

In recent years, Vision-Language Models (VLMs) \citep{radford2021learning,alayrac2022flamingo,li2023blip,dai2023instructblip,chen2023minigpt} as the technical basis of visual tasks such as Visual Question and Answer (VQA), have significantly promoted the significant development of application areas such as embodied intelligent interaction and autonomous driving scene understanding by virtue of their excellent visual perception and semantic reasoning capabilities. At the same time, with the technical migration of Large Language Models (LLMs) \citep{raffel2020exploring,wei2021finetuned,ouyang2022training,touvron2023llama,achiam2023gpt} to the multimodal domain, the resulting Multimodal Large Language Models (MLLMs) \citep{zhang2023llama,liu2023visual,girdhar2023imagebind,chen2023videollm} further break through the modal boundaries, not only realising cross-modal communication of text, image, video, and audio, but also performing higher-order tasks such as visual question and answer, visual localisation, and cross-modal retrieval. It is noteworthy that the emerging 3D MLLMs \citep{hong20233d,xu2024pointllm,guo2023point,qi2024gpt4point,mildenhall2021nerf} break through the 2D visual limitations and lay the technical foundation for semantic reconstruction and dynamic reasoning of 3D scenes through point cloud-text cross-modal mapping, marking an important extension of VLMs to the spatial intelligent cognition domain.

Although 3D MLLM facilitates semantic reasoning in scenes, its emphasis on geometric features necessitates the use of high-precision 3D representation approaches. In this context, as a new 3D representation paradigm, Neural Radiation Field (NeRF) steadily overcomes the limits of classic image-based explicit 3D data. On the one hand, based on the differentiable body rendering framework, it is capable of reconstructing high-fidelity 3D geometry from sparse 2D observation data, which provides innovative solutions for fields such as virtual reality; on the other hand, the continuous implicit representation feature enables the model to store only the parameters of the multilayer perceptual machine (MLP), which enables the synthesis of high-resolution views from arbitrary viewpoints, and breaks through the physical limitation of discrete data storage. Currently, academics have built a standardised evaluation system with the open-source NeRF dataset \citep{ramirez2023deep,amaduzzi2024llana}, which provides quantitative benchmarks for iterating 3D representation strategies. Scholars are exploring the deep fusion of NeRF with 3D MLLMs to accomplish direct processing of NeRF weights and develop a general-purpose NeRF assistant that can interpret and generate natural language \citep{qi2017pointnet++}.

As the integration of 3D MLLMs and NeRF representation methods becomes more widely used in practical applications, real-time understanding and rapid response capabilities for 3D scenes become increasingly critical. However, current NeRF-based approaches have inherent constraints when processing large-scale point cloud data. On the one hand, NeRF's performance is heavily dependent on the quality of the input data, and sparse or noisy point clouds will significantly reduce reconstruction quality. On the other hand, traditional feature extraction methods rely on fixed computational modes, making it difficult to strike a balance between expressiveness and computational efficiency. Although variants such as SparseNeRF \citep{wang2023sparsenerf} and combining Transformer attention \citep{wang2022attention} improve the modelling ability under sparse data to some extent, the nature of the sparse observation limits the complex geometry and textures from being accurately recovered, and data noise complicates optimisation.

To address the challenges raised above, this study offers a dynamic resolution multi-scale voxelized point cloud feature extraction method (DR-MSV) for efficiently constructing hierarchical representations of 3D scenes with varying fine-grain sizes. The method adaptively adjusts the voxel resolution based on multidimensional geometric and structural metrics, significantly reducing computational overhead while preserving the details of the objects in the scene, and providing dependable technical support for achieving speed and geometric fidelity objectives in complex scenes. Furthermore, to address the limitations of the traditional approach to model meta-encoder output feature aggregation, this paper introduces the Token-Level Adaptive Pooling for Lightweight Meta-Embedding (TAP-LME) mechanism, which improves the extraction and representation of more fine-grained features through the use of a token-level attentional weighting strategy.

Our contributions can be summarised as follows. 
\begin{itemize}
\item Proposed a high-performance point cloud voxelization method that increases feature extraction speed while maintaining reconstruction accuracy.
\end{itemize}
\begin{itemize}
\item Demonstrates, following evaluation on both short and detailed headings, that it is possible to add a lightweight token-level pooling architecture into the LLaNA model, which can significantly increase the model's performance on specific tasks.
\end{itemize}

\section{Related works}
\paragraph{Language-based 3D scene understanding}
Early research on 3D scene understanding has focused on semantic segmentation and instance segmentation tasks, which mainly rely on local neighbourhood feature aggregation mechanisms, such as point cloud convolution \citep{qi2017pointnet,qi2017pointnet++,wang2019dynamic} and voxel pooling \citep{graham2017submanifold,choy20194d}. However, such mechanisms are deficient in capturing long-distance spatial relations and cross-modal semantic links, which limits the generalisation performance of the model in dealing with occlusion, scale changes, and dynamic scenes. As research continues to evolve, 3D scene understanding is undergoing a gradual shift from a traditional perception-driven model to an interactive understanding model with natural language as a link, which has given rise to such mechanisms as 3D Visual Grounding (3DVG)\citep{chen2020scanrefer,achlioptas2020referit3d,chen2021mvt,zhang2023multi3drefer,chen2022language}, 3D Question Answering (3DQA)\citep{chen2021scan2cap,yuan2022x,chen2023end,chen2024vote2cap}, and 3D Dense Captioning \citep{azuma2022scanqa,parelli2023clip,ma2022sqa3d,ye20223d,zhao2022toward}. Specifically, 3DVG aims to accurately locate target objects in a 3D scene based on natural language queries, while 3DQA focuses on the model to complete Q\&A reasoning about scene perception based on natural language, and 3D Dense Captioning requires the model to achieve accurate location and semantic annotation of multiple targets in the scene. The proposal of such tasks marks the deep migration of 3D visual understanding from a passive perception paradigm to an active reasoning paradigm, which provides a technical extension for more intelligent practical applications.

\paragraph{Zero-sample LLM/VLM for 3D scene understanding}
Researchers have gradually applied LLMs \citep{touvron2023llama,achiam2023gpt,chiang2023vicuna,touvron2023llama,chowdhery2023palm} to 3D scene understanding tasks as a result of their excellent performance in complex environment reasoning and natural language interaction, pushing the field towards a language-driven active understanding paradigm. Among these, PointLLM \citep{xu2024pointllm} encodes geometric aspects of point clouds to offer structured input for the semantic representation of LLMs. Another line of study \citep{girdhar2023imagebind,guo2023point,han2023imagebind,jiao2024unlocking} uses a joint embedding space of 3D point clouds, pictures, and text to feed 3D modalities into LLM for representation. These models are competitive in comprehending pictures and 3D data, but they fall short in capturing object appearance and geometry. To address this limitation, subsequent research \citep{xue2023ulip,xue2024ulip,kerr2023lerf,yu2022point,pang2022masked} has improved the ability to model object appearance by introducing image texturing or reconstruction mechanisms, as well as to represent geometry more efficiently with advanced point cloud encoders, thereby improving fine-grained comprehension of language-driven tasks in 3D scenes.

\paragraph{Neural radiance fields}
Neural Radiance Fields (NeRF) \citep{mildenhall2021nerf} has gradually expanded from novel view synthesis to a variety of application scenarios, such as generative content modelling, robot perception, etc., due to its ability to achieve the representation of the mapping between spatial coordinates and colours/densities based on MLP. To address the efficiency bottleneck of traditional MLP, studies \citep{chen2022tensorf,sun2022direct,fridovich2022plenoxels,muller2022instant} switched to explicit scene representation and neural field fusion architectures, which dramatically improved the spatio-temporal efficiency of training inference. NeRF's interaction with natural language has evolved into an emerging research hotspot, encompassing scene/object generation based on textual cues \citep{seo2023ditto,metzer2023latent,jo2023cg,sen2023hyp}, scene editing (e.g., appearance and shape modification \citep{wang2023nerf,sun2024nerfeditor} and object addition and deletion \citep{bai2023componerf,mirzaei2023reference}), and neural field construction for language feature prediction \citep{kerr2023lerf,kobayashi2022decomposing}. Existing work includes LERF \citep{kerr2023lerf}, which performs text-based field searches by predicting local linguistic features, and Ballerini et al.\citep{ballerini2024connecting}, who investigate the mapping between NeRF weights and CLIP embeddings for retrieval tasks. Furthermore, meta-network research in deep learning provides the theoretical foundation for this work, with early work focussing on predicting performance from network weights \citep{unterthiner2020predicting,schurholt2021self,knyazev2021parameter}, and more recent studies exploring the extraction of feature vectors that can be used for downstream tasks directly from the network parameters of implicit neural representations \citep{ramirez2023deep,de2023deep,cardace2023neural,navon2023equivariant}. 

\section{Methods}
\begin{figure}[h!]
  \centering
  \includegraphics[width=1\textwidth]{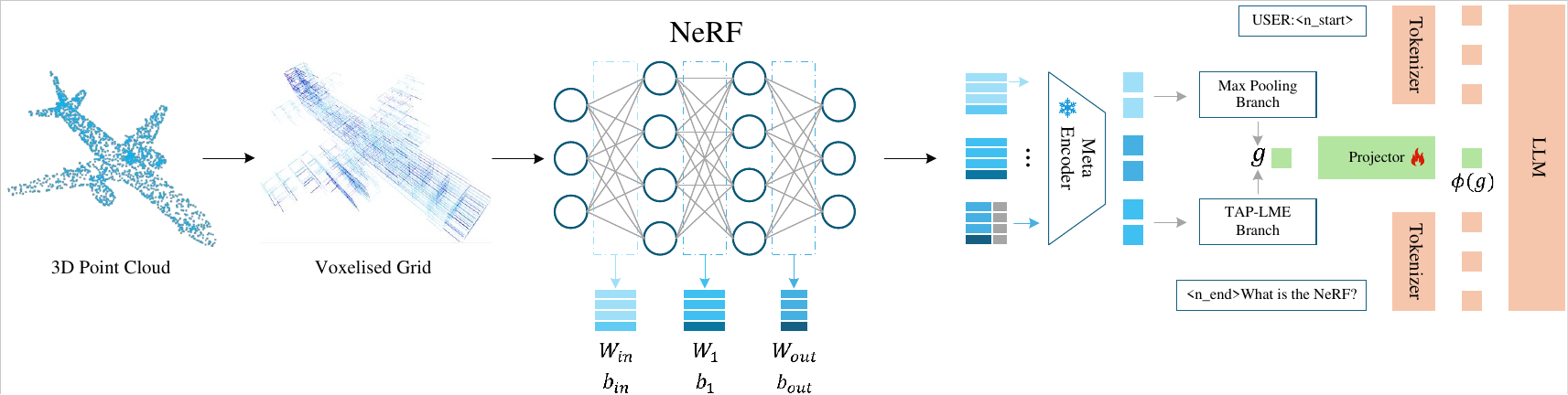}
  \caption{ \textbf{Overall architecture of NeuroVoxel-LM.} }
  \label{2}
\end{figure}
This section discusses the proposed Neural Voxel-based Multimodal Language Model (NeuroVoxel-LM) (shown in Fig. 1). Subsection 1 introduces the Dynamic Resolution Multiscale Voxelisation (DR-MSV) method for optimizing the 3D point cloud feature extraction strategy, aiming to achieve guaranteed reconstruction accuracy while improving the speed of point cloud extraction. Subsection 2 then introduces a Token-Level Adaptive Pooling for Lightweight Meta-Embedding (TAP-LME) mechanism to replace the maximal pooling phase in the original architecture in order to optimise the balance between computational efficiency and expressiveness.
\subsection{Dynamic resolution multi-scale voxelisation}
To increase geometric appearance fidelity and computational efficiency in 3D point cloud representation, this work offers a dynamic resolution multi-scale voxelisation (DR-MSV) approach (as shown in Fig. 2). The approach assesses voxel complexity using multiple geometric and structural indicators before constructing a multilevel voxel pyramid based on a layer-by-layer judgement mechanism to enable efficient point cloud scene modelling and adaptive resolution control.

\begin{figure}[h!]
  \centering
  \includegraphics[width=1\textwidth]{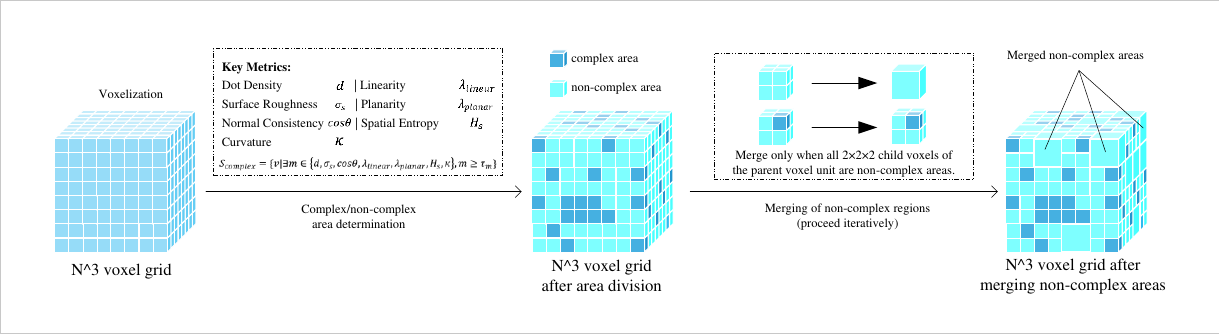}
  \caption{ \textbf{DR-MSV methods.} After voxel processing of the input 3D point cloud data, the complexity of voxel units is determined based on key metrics, followed by iterative merging of non-complex regions.}
  \label{3}
\end{figure}

During the initialisation stage, the input point cloud is divided into a fixed-size voxel grid (e.g., 16 × 16 × 16) to create a fine-grained geometric representation. Then, each voxel cell in the initialised voxel network is determined whether it is a complex region or not based on the complexity assessment criteria consisting of the following six indicators (as shown in Algorithm 1 in the Appendix).
\paragraph{Dot density $d$} Measures the number of points within a voxel and is used to assess the richness of local geometric detail. The region is typically more complex as the density increases.
    \begin{equation}
        d = \frac{n}{V_{\text{voxel}}} \label{eq:dot_density}
    \end{equation}
\paragraph{Surface roughness $\sigma_s$}  The degree of undulation of the local surface is described by the mean square error from the point to the fitted plane, and complicated structures such as curved surfaces or fissures can be identified accurately.
    \begin{equation}
        \sigma_s = \sqrt{\frac{1}{n} \sum_{i=1}^{n} (p_i - \pi)^2} 
    \end{equation}
\paragraph{Normal coherence $\cos\theta$} A measure of how the normal vector changes direction within a voxel, which is used to capture structural boundaries and geometric mutation zones.
    \begin{equation}
        \cos\theta = \frac{1}{n} \sum_{i=1}^{n} \mathbf{n}_i \cdot \mathbf{n}_{\text{avg}} 
    \end{equation}
\paragraph{Structural principal component indicators} Characterising the linear ($\lambda_{\text{linear}}$) and planar ($\lambda_{\text{planar}}$) features of a point collection using PCA aids in the identification of structural regions with diverse primary morphologies.
        \begin{equation}
            \lambda_{\text{linear}} = (\lambda_1 - \lambda_2) / \lambda_1 
        \end{equation}
        \begin{equation}
            \lambda_{\text{planar}} = (\lambda_2 - \lambda_3) / \lambda_1 
        \end{equation}
\paragraph{Spatial distribution entropy $H_s$} Calculates voxel-wise entropy to measure spatial point dispersion. High entropy values typically indicate sparse or non-uniform distributions, which are common in fractured or irregular structures.
        \begin{equation}
            H_s = - \sum p_b \log p_b
        \end{equation}
\paragraph{Curvature estimation $\kappa$} Reflects the curvature of local surfaces and can recognise locations with high geometric detail, such as edges and corner points.
        \begin{equation}
    \kappa = \frac{\lambda_3}{\lambda_1 + \lambda_2 + \lambda_3}
        \end{equation}
Thus, the sets of complex regions and non-complex regions are defined as:
\begin{equation}
    S_{\text{complex}} = \{ \forall m \in \{d, \sigma_s, \cos\theta, \lambda_{\text{linear}}, \lambda_{\text{planar}}, H_s, \kappa\}, m \ge \tau_m \} \label{eq:complex_set}
\end{equation}
\begin{equation}
    S_{\text{non\_complex}} = \{ \forall m \in \{d, \sigma_s, \cos\theta, \lambda_{\text{linear}}, \lambda_{\text{planar}}, H_s, \kappa\}, m < \tau_m \} \label{eq:non_complex_set}
\end{equation}
Note that in order to maintain the generality and rigor of complexity determination, manually adjusted fixed thresholds are eschewed in favor of a data-driven percentile-based thresholding technique. The threshold for each complexity indicator is established automatically using the training data's empirical distribution. Specifically, a voxel is termed complex if its metric exceeds the 75th percentile of the related measure across the scene. This adaptive method enables the model to generalise across circumstances without requiring hard-coded assumptions.

After determining the region at the maximum resolution, we enter the iterative merging step (As shown in Algorithm 2 in the Appendix). A 2 × 2 × 2 voxel block is the smallest merging unit based on the current voxel resolution. A region is regarded as structurally simple and allowed to be merged only if all 8 child voxels within the block are non-complex regions. 8 non-complex region child voxels are united to form a larger-volume parent voxel that remains non-complex. The merging is done iteratively, layer by layer, and each merging round gradually merges the smaller non-complex voxels into the larger voxels until all of the voxel blocks in the point cloud scene do not satisfy the merging condition, resulting in the formation of a voxel pyramid made up of voxel blocks of varying sizes.
\subsection{Token-Level Adaptive Pooling for Lightweight Meta-Embedding}
Although the nf2vec-based meta-encoder can efficiently extract token features from NeRF weights, its global feature representation remains based on a simple max-pooling process. It is unable to describe the semantic contribution of each token when confronted with complicated geometric structures and texture information, which can easily result in the loss of critical information. To that end, a new pooling mechanism, Token-Level Adaptive Pooling for Lightweight Meta-Embedding (as shown in Fig. 3), is proposed. By assigning learnable attentional weights to each token, TAP-LME improves the understanding of fine-grained meta-embedding by keeping it lightweight.

\begin{figure}[h!]
  \centering
  \includegraphics[width=1\textwidth]{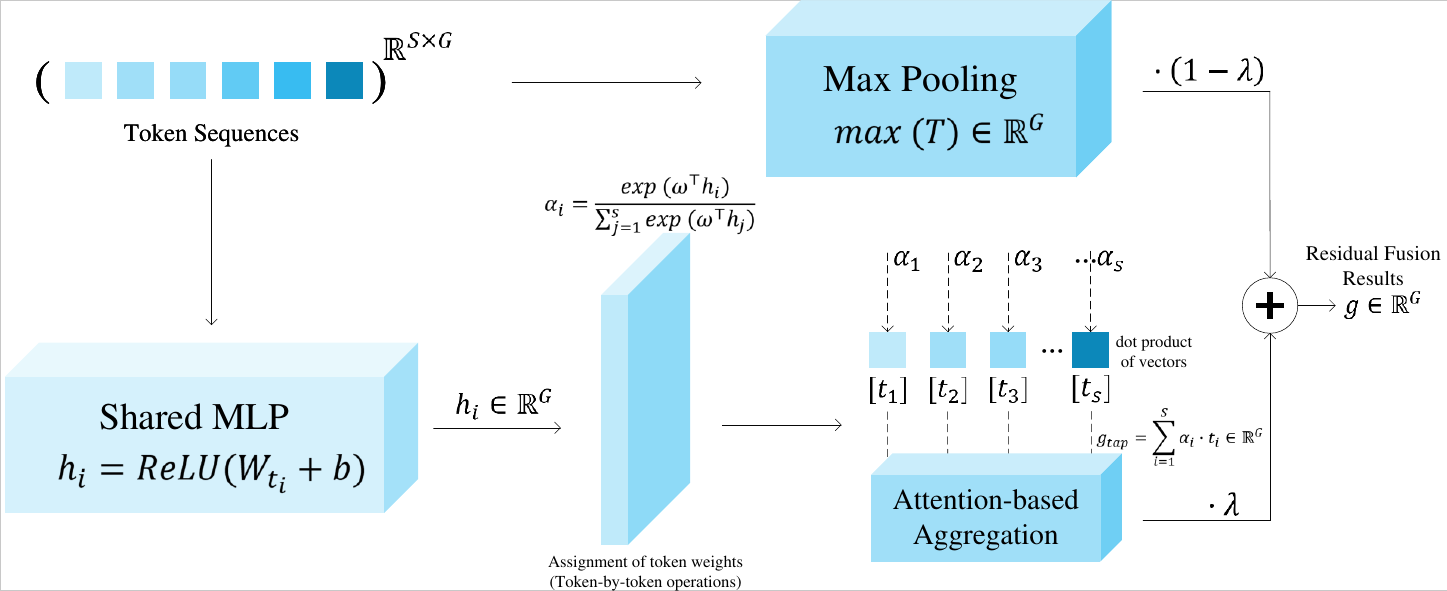}
  \caption{ \textbf{Architecture of TAP-LME.} The model processes a token sequence $\mathbf{T}$ through two branches: one applies max pooling to generate a global representation, while the other uses a shared MLP and attention mechanism to produce an adaptive representation $g_{tap}$. These are then fused via a residual connection with a learnable parameter $\lambda$ to yield the final output $g$.}
  \label{1}
\end{figure}

\paragraph{Token-level attention pooling}
The encoder's token sequence can be represented by the matrix $\mathbf{T} = [t_1, t_2,..., t_s] \in \mathbb{R}^{S \times G}$, where each token vector $t_i \in \mathbb{R}^{G}$ represents semantic characteristics derived from distinct substructures of the NeRF weight structure.

In the original architecture, all tokens are routed straight into the maximum pooling layer, resulting in a global representation. However, this technique ignores the relative value of distinct tokens at the level and hence misses potentially relevant information. As a result, this paper introduces a lightweight attentional weighting algorithm that assigns adaptive importance weights to each token, allowing the overall feature to better reflect the important information of the original NeRF.

Each token $t_i$ is first fed into a shared MLP layer for nonlinear mapping, and then the attention weights are computed by the learnable parameter $\mathbf{w}$: 
\begin{equation}
    \alpha_i = \frac{\exp(\mathbf{w}^\top \text{ReLU}(\mathbf{W} t_i + b))}{\sum_{j=1}^{s} \exp(\mathbf{w}^\top \text{ReLU}(\mathbf{W} t_j + b))} \label{eq:attention_weight}
\end{equation}
where $\mathbf{W} \in \mathbb{R}^{(G \times G)}$, $\mathbf{w} \in \mathbb{R}^{G}$, $b \in \mathbb{R}^{G}$ are shared learnable parameters, and ReLU denotes the element-by-element nonlinear activation function used to introduce nonlinear discriminative power.

This approach can effectively demonstrate the relative relevance of tokens, causing the model to pay greater attention to tokens with key geometric structures or semantic hints, resulting in robust and discriminative global feature extraction in many contexts.

The attention weights $\alpha_i$ are then utilized to weight and fuse all tokens, resulting in an adaptive global representation vector $g_{tap}$:
\begin{equation}
    g_{tap} = \sum_{i=1}^{s} \alpha_i \cdot t_i \label{eq:gtap}
\end{equation}
\paragraph{Residual fusion with maximum pooling}
Based on the original maximum pooling, the TAP-LME mechanism is introduced to enhance the model's dual modelling capability of local details and overall structure by means of residual fusion. Specifically, the adaptive pooling result $g_{tap}$ acquired by attentional weighting is linearly fused with the global vector $\max(\mathbf{T})$ produced by traditional maximal pooling:
\begin{equation}
    g = \lambda \cdot g_{tap} + (1 - \lambda) \cdot \max(\mathbf{T}) \label{eq:g}
\end{equation}
where the fusion factor $\lambda \in [0, 1]$ is used to control the relative proportions of the two. To make feature fusion more flexible, the fusion coefficient $\lambda$ is designed as a trainable scalar. As a result, the model is able to adaptively decide the resource allocation between attention pooling and maximum pooling during training, eliminating the difficulty of artificially adjusting $\lambda$ and ensuring the rigour of the algorithm implementation.

\section{Experiments}
This section validates the effectiveness of DR-MSV with TAP-LME in the NeuroVoxel-LM framework through systematic experiments. The experiments are designed around two core objectives:
\paragraph{Algorithm Performance Comparison} DR-MSV is compared to the baseline approach, Fixed-Resolution Voxelisation (FRV), to quantify its advantages in geometric reconstruction accuracy and computing efficiency. Meanwhile, the TAP-LME mechanism is contrasted to the standard maximum pooling method to see how effective it is in improving fine-grained feature expression capabilities through attention weighting and residual fusion.

\paragraph{Module Contribution Analysis} The dynamic merging approach of DR-MSV is removed and simplified to binary classification, and many TAP-LME variants are built using various pooling algorithms to determine the impact of each submodule on overall performance.
\subsection{Comparative Experiments}
\subsubsection{Dynamic resolution multi-scale voxelisation}
\paragraph{Hardware Environment} The training and comparative evaluation of the DR-MSV method with the old FRV method was done on 1 NVIDIA L20 (with 48GB VRAM). It takes 1-2 days of training time to complete these phases.
\paragraph{Dataset} We conducted experiments on the ShapeNet \citep{chang2015shapenet} benchmark dataset, a large-scale, richly annotated library of 3D CAD models covering a wide range of object classes. It serves as a recognised benchmark in the fields of computer vision and graphics, providing standardised training, validation and test data for tasks such as 3D shape understanding, part segmentation and reconstruction.
\paragraph{Metrics} In this study, model performance was assessed in terms of training efficiency and reconstruction accuracy. 
For training efficiency, we recorded and calculated: 
Total Training Time (TT), 
Single-batch Processing Time (BT), 
Data Preparation Time (DP), 
Model Fitting Time (MF), 
and shape processing speed (SP). For reconstruction accuracy, an international common index system was used: the Chamfer Distance to assess the quality of point cloud reconstruction, the F1 score (Point Cloud), and the Geometric IoU to measure the degree of spatial overlap. For voxelization results, Accuracy, Precision, Recall, F1 Score (Voxel), and voxel IoU are further evaluated to examine the classification performance comprehensively.
\paragraph{Baseline} The baseline method employs a fixed-resolution voxelization strategy to predict the voxel mesh directly from the point cloud via the SIREN network, relying on Focal Loss optimisation. However, its fixed-resolution characteristic leads to limited computational efficiency, especially when processing high-resolution voxels, which significantly reduces the feature extraction speed of the point cloud and makes it difficult to meet the real-time demand.

The proposed DR-MSV approach greatly surpasses the FPV baseline in terms of training efficiency and reconstruction accuracy. As indicated in Table 1, DR-MSV reduces total training time by over 35\%. In terms of reconstruction quality (Table 2), DR-MSV improves significantly across all measures, particularly Chamfer Distance and voxel IoU, demonstrating its ability to capture fine-grained spatial information while accelerating computation.
\subsubsection{Token-Level Adaptive Pooling for Lightweight Meta-Embedding}
\paragraph{Hardware Environment} The TAP-LME method was trained and compared to the traditional maximum pooling method using four NVIDIA A800s (each with 80GB VRAM). Completing these phases requires approximately one day of training time.
\paragraph{Dataset} Experiments were carried out using the ShapeNeRF-Text \citep{amaduzzi2024shapenerftext} benchmark dataset, which combines NeRF modelling and text annotation and includes NeRF representations of ShapeNet objects as well as a diverse set of text descriptions and Q\&A pairings. It sets a standard for NeRF-language comprehension activities, including NeRF subtitle creation and Q\&A.
\paragraph{Metrics} Given the input NeRF, the NeRF assistant's performance was evaluated on two linguistic tasks: short headings and detailed headings. The quality of the generated text was assessed using learning-based metrics based on Sentence-BERT, SimCSE, and n-gram metrics such as BLEU, ROUGE, and METEOR.
\paragraph{Baseline} The base model LLaNA \citep{amaduzzi2024llana} is a large language and NeRF assistant that pioneered the application of large language models to neural radiance. It uses NeRF's MLP weights directly rather than using traditional rendering or 3D reconstruction to understand the objects represented by NeRF, allowing it to enable a wide range of NeRF-language interface activities such as NeRF subtitle production and question answering.

Based on the ShapeNeRF-Text dataset, LLaNA-7b serves as the baseline, and all TAP-LME variations outperform in both short title and detailed title generating tasks. The TAP-Res (Learnt) version, in particular, achieves the highest scores in S-BERT and SimCSE indicators, indicating that the TAP mechanism effectively improves the quality of the generated text through adaptive fusion learning. Although the improvement in indicators is minimal, the overall trend suggests that TAP-LME improves NeRF language understanding.
\subsection{Ablation Experiments}
\subsubsection{Dynamic resolution multi-scale voxelization}
\begin{table}[htbp]
  \caption{Analysis of Point Cloud Voxelization Efficiency Ablation}
  \label{tab:training-metrics}
  \centering
  \begin{tabular}{lccccc}
    \toprule
    Model & TT (h) & DP (h) & MF (s) & BT (s) & SP (shapes/s) \\
    \midrule
    FPV   & 19.87 & 4.67 & 15.20 & 19.94 & 0.81 \\
    DR-MSV & 12.24 & 2.56 & 9.68 & 12.28 & 1.31 \\
    DR-SV-MC & 11.87 & 2.28 & 9.59 & 11.91 & 1.36 \\
    DR-MSV-BC & 10.23 & 1.49 & 8.74 & 10.27 & 1.57 \\
    \bottomrule
  \end{tabular}
\end{table}
\begin{table}[htbp]
  \caption{Analysis of Point Cloud Voxelization Accuracy Ablation}
  \label{tab:voxel-accuracy-ablation}
  \centering
  \begin{tabular}{lcccccccc}
    \toprule
    Model & CD & F1(PC) & GeoIoU & Acc & Prec & Rec & F1(Vox) & VoxIoU \\
    \midrule
    FPV         & 0.195 & 0.62 & 0.47 & 0.69 & 0.64 & 0.61 & 0.60 & 0.45 \\
    \textbf{DR-MSV}      & \textbf{0.150} & \textbf{0.74} & \textbf{0.56} & \textbf{0.74} & \textbf{0.72} & \textbf{0.71} & \textbf{0.73} & \textbf{0.56} \\
    DR-SV-MC    & 0.188 & 0.64 & 0.49 & 0.71 & 0.66 & 0.63 & 0.62 & 0.47 \\
    DR-MSV-BC   & 0.182 & 0.65 & 0.50 & 0.72 & 0.67 & 0.64 & 0.63 & 0.48 \\
    \bottomrule
  \end{tabular}
\end{table}
In ablation experiments (as shown in Tables 1 and 2), we examine different strategies for voxelising point clouds. Our FPV baseline employs a typical single-resolution grid with straightforward binary placeholder labelling. Our main method, DR-MSV, improves on this approach by including extensive voxel analyses, combining multiresolution data, and predicting multiclass complexity labels. To better appreciate the impact of these advancements, we remove essential components: the Dynamic Resolution with Single-Scale Voxels and Multi-Class (DR-SV-MC) eliminates multiscale data fusion, while the Dynamic Resolution with Multi-Scale Voxels and Binary Classification (DR-MSV-BC) reduces the prediction task to binary classification. The hardware environment, dataset, and metrics were used in the same way as in the comparison experiments.

The ablation results validate the role of each DR-MSV component. Removing multiscale fusion (DR-SV-MC) or limiting the output to binary classification (DR-MSV-BC) reduces accuracy slightly but increases training efficiency due to lower computing complexity. DR-SV-MC and DR-MSV-BC perform faster shape processing than the entire DR-MSV model, but have poorer voxel-level F1 and IoU scores. These findings show that simplifying the processing pipeline increases speed at the expense of reconstruction accuracy, emphasising the importance of multiscale integration and multiclass supervision for exact voxel prediction.
\subsubsection{Token-Level Adaptive Pooling for Lightweight Meta-Embedding}
To test the efficiency of the Token-Level Adaptive Pooling (TAP-LME) module presented in this research in improving global feature representation, we developed a series of ablation tests. There are four models: a baseline model (Baseline) that only uses maximal pooling, a TAP-only version that only uses attention-weighted pooling, a TAP-Res model that combines maximal pooling with TAP and is fused at a fixed ratio (($\alpha$)=0.5), and a TAP-Res (learnt) model with learnable fusion coefficients. In addition, a TAP-weight-only version with attention weights set to the mean was introduced to validate the attention mechanism. The actual contribution and benefits of the TAP-LME mechanism are carefully assessed by comparing each version's performance in terms of accuracy and other factors. The results of the ablation experiment results are shown in Table 3 and 4.

Ablation studies show that the components of the TAP-LME module contribute significantly to the generation of NeRF short and detailed titles. Among them, the TAP-Res (Learnt) model performs best, leading in semantic similarity indicators such as S-BERT and SimCSE and generation quality indicators such as ROUGE-L and METEOR, indicating that the strategy of combining attention weighting with maximum pooling and dynamically fusing through learnable coefficients effectively improves the global feature representation. Hybrid pooling is better than a single strategy: TAP-only (attention only) is better than the baseline, but not as good as TAP-Res; the TAP-Res model with fixed ratio fusion ($\alpha$=0.5) performs better, proving that fusing the two pooling methods is more advantageous. In contrast, TAP-Res (Learnt) further optimizes the performance by adaptively adjusting the fusion weights. In addition, the TAP-Weight-only model, which sets the attention weights to the mean, performs poorly, emphasizing the importance of the true attention mechanism in identifying key information. In summary, the ablation experiments strongly demonstrate the effectiveness of the TAP-LME module, especially its mixed pooling mechanism with learnable fusion coefficients.
\begin{table}[htbp]
  \caption{Experimental results of NeRF short headings on ShapeNeRF-Text}
  \label{tab:nerf-shapenerf-text}
  \centering
  \begin{tabular}{lcccccc}
    \toprule
    Model & Modality & S-BERT & SimCSE & BLEU-1 & ROUGE-L & METEOR \\
    \midrule
    LLaVA-vicuna-7b    & Image (FV)    & 54.31  & 56.28  & 10.08  & 14.71   & 14.53  \\
    LLaVA-vicuna-7b    & Image (BV)    & 51.75  & 52.29  & 8.13   & 13.96   & 14.18  \\
    PointLLM-7b        & Point cloud   & 43.40  & 44.50  & 8.53   & 11.64   & 9.97   \\
    GPT4Point-Opt-2.7B & Point cloud   & 43.15  & 42.22  & 12.02  & 18.73   & 13.69  \\
    3D-LLM               & Mesh + MV         & 56.07           & 52.13           & 15.94           & 20.71            & 15.22           \\
    LLaNA-7b (Baseline) & NeRF & 68.63 & 70.54 & 20.64 & 28.33 & 31.76 \\
    TAP-only & NeRF & 68.85 & 70.50 & \textbf{20.79} & 28.21 & 31.73 \\
    TAP-Res ($\alpha=0.5$) & NeRF & 68.89 & 70.70 & 20.54 & 28.36 & 31.67 \\
    TAP-Res (Learnt) & NeRF & \textbf{69.10} & \textbf{70.76} & 20.75 & \textbf{28.42} & \textbf{31.83} \\
    TAP-Weight-only & NeRF & 68.75 & 70.30 & 20.67 & 28.17 & 31.59 \\
    \bottomrule
  \end{tabular}
  \smallskip
  \textit{Best results are in \textbf{bold}. (FV: front-view, BV: back-view, MV: multi-view)}
\end{table}
\begin{table}[htbp]
  \caption{Experimental results of NeRF detailed headings on ShapeNeRF-Text}
  \label{tab:nerf-detailed-headings}
  \centering
  \begin{tabular}{lcccccc}
    \toprule
    Model & Modality & S-BERT & SimCSE & BLEU-1 & ROUGE-L & METEOR \\
    \midrule
    LLaVA-vicuna-7b  & Image (FV) & 71.79 & 71.96 & 25.79 & 34.04 & 34.86 \\
    LLaVA-vicuna-7b  & Image (BV) & 70.88 & 70.93 & 25.17 & 33.30 & 34.22 \\
    PointLLM-7b      & Point cloud & 74.70 & 74.40 & 36.81 & \textbf{44.41} & \textbf{39.76} \\
    GPT4Point-Opt-2.7b & Point cloud & 27.62 & 31.41 & 6.26 & 9.38 & 5.41 \\
    3D-LLM & Mesh + MV & 60.00 & 53.91 & 1.58 & 14.40 & 5.28 \\
    LLaNA-7b (Baseline) & NeRF & 77.43 & 79.81 & 41.32 & 36.18 & 32.39 \\
    TAP-only & NeRF & 78.12 & 79.87 & 41.47 & 36.08 & 32.13 \\
    TAP-Res ($\alpha=0.5$) & NeRF & 78.24 & 79.73 & 41.38 & 36.27 & 32.44 \\
    TAP-Res (Learnt) & NeRF & \textbf{78.33} & \textbf{79.91} & \textbf{41.62} & 36.39 & 32.49 \\
    TAP-Weight-only & NeRF & 77.96 & 79.59 & 41.54 & 36.11 & 32.22 \\
    \bottomrule
  \end{tabular}
  \smallskip
  \textit{Best results are in \textbf{bold}. (FV: front-view, BV: back-view, MV: multi-view)}
\end{table}
\section{Conclusion}
In order to solve the problem that the existing 3D language models have slow feature extraction and low accuracy when processing sparse, large-scale point clouds, this paper proposes the LLaNA-DRVNet framework. This framework innovatively combines Neural Radiance Field (NeRF) with Dynamic Resolution Voxelization (DR-MSV) and Lightweight Meta-Embedding (TAP-LME) technology. Experiments have shown that DR-MSV effectively improves the speed of point cloud feature extraction and maintains the accuracy of geometric reconstruction, while TAP-LME improves the semantic understanding of NeRF weights through attention weighting and residual fusion. NeuroVoxel-LM provides a new solution for achieving more efficient and accurate language-driven 3D scene perception.
\bibliographystyle{plain} 
\bibliography{main}

\newpage

\end{document}